%% file: main.tex
\documentclass[11pt]{article}
\usepackage[T1]{fontenc}
\usepackage{lmodern,amsmath,amssymb,amsthm,booktabs,graphicx,microtype}
\usepackage[margin=1in]{geometry}
\usepackage[colorlinks=true,linkcolor=blue,citecolor=blue,urlcolor=blue]{hyperref}
\hypersetup{pdftitle={Budgeted Proper Loss Design for Uncertain Decision Thresholds},pdfauthor={Ankit Sharma}}
\newtheorem{theorem}{Theorem}
\newtheorem{proposition}[theorem]{Proposition}

\newcommand{\E}{\mathbb{E}}
\newcommand{\1}{\mathbf{1}}
\newcommand{\esssup}{\operatorname*{ess\,sup}}
\title{Budgeted Proper Loss Design for Uncertain Decision Thresholds}
\author{Ankit Sharma\\[3pt]
\small International Institute of Information Technology Bangalore\\
\small Bengaluru, India\\
\small \texttt{ankitankitsharm.eml36@iiitb.net}}
\date{September 2026}
\begin{document}
\maketitle
\begin{abstract}
A binary probability estimate may be used by several decision makers with different error costs. Proper scoring rules can emphasize selected probability thresholds, but a single average threshold distribution need not give a good worst-case regret guarantee. We study the choice of a proper loss when the threshold distribution belongs to a finite specified family. Among integrable loss weights with a positive pointwise floor and a fixed total mass, we derive the weight that minimizes the sharp, distribution-free multiplier from surrogate regret to threshold-averaged decision regret. The solution clips a rescaled upper envelope of the threshold densities at the prescribed floor; its normalization is found by a one-dimensional root search. We give proofs of sharpness, optimality, and stability under density misspecification, as well as a limitation for atomic threshold laws. An exact histogram implementation and two synthetic experiments across 20 seeds verify the construction. The optimized multiplier improves over uniform and average-density weights, while fitted-model performance does not improve uniformly. The result is a loss-design guarantee under an explicit budget, rather than a general claim of predictive superiority.
\end{abstract}

\section{Introduction}
A model that estimates a binary event probability is often shared across decisions. A prediction may determine whether to allocate a resource, flag an item, or accept an opportunity. When the relative costs of false positives and false negatives differ across these uses, the same probability is compared with different thresholds. Choosing one training loss then raises a practical question: which parts of the probability interval should receive additional emphasis?

Threshold emphasis is not a new loss principle. Binary proper scoring rules admit a representation as weighted mixtures of cost-sensitive decisions \cite{buja,reid2010,reid2009}. Tailoring these weights to a relevant threshold is established practice \cite{buja}, and weighted Brier scores explicitly average over threshold preferences \cite{zhu}. This paper instead asks a constrained design question. Suppose several threshold laws are plausible, no mixing probabilities among them are specified, and the training loss must retain a uniform probability-estimation floor. How should a fixed amount of additional weight be allocated to minimize a worst-case regret multiplier?

The resulting problem has a simple exact solution. Take the pointwise maximum of the threshold densities, divide it by a scalar, and clip it from below at the required floor. Choose the scalar so that the resulting weight uses the available mass. We call the induced scoring rule the \emph{envelope loss}. The name denotes this particular design rule within the existing class of proper losses; it does not denote a newly discovered representation of scoring rules.

The principal result is the solution of this budgeted design problem together with an exact characterization of the multiplier being optimized. The analysis also quantifies sensitivity to an inaccurate threshold family and explains why a locally bounded weight cannot give the same linear guarantee for a fixed atomic threshold. A numerical study distinguishes this guarantee from the separate question of which fitted model performs best.

\section{Setting and relation to existing work}
Let $Y\in\{0,1\}$, let $X$ be a feature vector, and write $\eta(x)=\Pr(Y=1\mid X=x)$. A prediction $q(x)$ takes values in $[0,1]$. For threshold $t\in(0,1)$, use action $a_q(t)=\1\{q\geq t\}$ and cost
\begin{equation}
 c_t(y,a)=t(1-y)a+(1-t)y(1-a).
\end{equation}
The Bayes action is $a_\eta(t)$. If $q$ and $\eta$ lead to different actions, the conditional excess cost is $|\eta-t|$.

We assume a specified finite family of bounded measurable densities $f_1,\ldots,f_K$ on $(0,1)$, each integrating to one. A density describes a distribution of thresholds, independently of $(X,Y)$, for one use of the model. Only this threshold law varies; the joint law of $(X,Y)$ is fixed. Define the conditional decision regret
\begin{equation}\label{eq:decision}
 D_f(\eta,q)=\int_{\min(\eta,q)}^{\max(\eta,q)} |t-\eta| f(t)\,dt,
 \qquad \mathcal R_f(q)=\E_X D_f(\eta(X),q(X)).
\end{equation}
The worst family regret is $\max_k\mathcal R_{f_k}(q)$. It is also the worst regret over the convex hull of the family, since expectation is linear in $f$. The maximum is outside the expectation: one law is used for the population, rather than selected separately for each example.

\paragraph{Proper loss representation.}
For an integrable weight $w\geq0$, define
\begin{equation}\label{eq:loss}
 \ell_w(y,q)=(1-y)\int_0^q t w(t)\,dt
            +y\int_q^1 (1-t)w(t)\,dt.
\end{equation}
This is the established integral representation used throughout the paper. Uniform weight $w=2$ gives $(y-q)^2$, the binary Brier loss. The expected loss $L_w(\eta,q)=\E_{Y\sim\mathrm{Bern}(\eta)}\ell_w(Y,q)$ has regret
\begin{equation}\label{eq:regret}
 D_w(\eta,q)=L_w(\eta,q)-L_w(\eta,\eta)
 =\int_{\min(\eta,q)}^{\max(\eta,q)}|t-\eta|w(t)\,dt.
\end{equation}
Write $\mathcal R_w(q)=\E_X D_w(\eta(X),q(X))$. In particular, $D_f$ is itself a proper-score regret. This identifies the design problem as a comparison of proper-score regrets, not a different notion of probability calibration.

\paragraph{Prior work and scope of the contribution.}
Buja, Stuetzle, and Shen \cite{buja} develop weight-based proper scoring rules and threshold tailoring. Reid and Williamson \cite{reid2010,reid2009} establish general relations among proper losses, links, divergences, and regret bounds. Their framework supplies the representation and calculus used here. Zhu, Zheng, and Chan \cite{zhu} study a weighted Brier score for decision-relevant evaluation. Flores et al.\ \cite{flores} examine cost-aware scoring under uncertainty in class balance, a different source of deployment variation. Painsky and Wornell \cite{painsky} give divergence-domination results for log loss under their regularity and convexity assumptions. These works make a broad claim that ``threshold-aware proper loss is new'' untenable. The contribution considered here is the explicit floor-and-mass constrained minimax allocation for a specified finite family, with a sharp coefficient and an exact implementation. We do not claim a new general scoring-rule theory or an exhaustive classification of robust losses.

\section{A sharp regret multiplier}
Fix constants $m>0$ and $B>0$, and let
\begin{equation}\label{eq:feasible}
 \mathcal W_{m,B}=\left\{w\in L^1(0,1): w\geq m\ \text{a.e.},\ \int_0^1w(t)\,dt=m+B\right\}.
\end{equation}
The floor preserves sensitivity over the full probability interval; $B$ is the extra mass that can be allocated. A mass constraint is essential: unrestricted multiplication of a loss by an arbitrarily large number would make its regret multiplier arbitrarily small without changing unregularized empirical minimizers.

Let $F(t)=\max_{1\leq k\leq K}f_k(t)$. For $w\in\mathcal W_{m,B}$, set
\begin{equation}\label{eq:constant}
 C(w)=\esssup_{t\in(0,1)}\frac{F(t)}{w(t)}.
\end{equation}
The essential supremum ignores changes on sets of Lebesgue measure zero.

\begin{theorem}[Properness and sharp domination]\label{thm:sharp}
For every $w\in\mathcal W_{m,B}$, $\ell_w$ is strictly proper and
\begin{equation}\label{eq:domination}
 D_w(\eta,q)\geq\frac m2(q-\eta)^2,
 \qquad \max_k\mathcal R_{f_k}(q)\leq C(w)\mathcal R_w(q).
\end{equation}
Moreover, $C(w)$ is the smallest nonnegative constant for which the second inequality holds for every joint law of $(X,Y)$ and every measurable predictor $q$.
\end{theorem}
\begin{proof}
The derivative of conditional risk is $(q-\eta)w(q)$ almost everywhere. Integration gives \eqref{eq:regret}. The lower bound follows from $w\geq m$ and is strictly positive when $q\neq\eta$. Since $f_k\leq C(w)w$ almost everywhere, integration in \eqref{eq:decision} followed by expectation proves domination.

For sharpness, let $0\leq c<C(w)$. As $K$ is finite, some $k$ has a positive-measure set on which $f_k(t)/w(t)>c$. Choose an interior common Lebesgue point $t_0$ of $f_k$ and $w$ in this set, with finite $w(t_0)\geq m$. Take a degenerate feature distribution with $\eta=t_0$ and prediction $q=t_0+h$. For $g=f_k$ or $w$, one-sided Lebesgue differentiation gives
\[
 \int_0^h u g(t_0+u)\,du=\tfrac12g(t_0)h^2+o(h^2).
\]
Indeed, the absolute remainder is at most $h\int_0^h|g(t_0+u)-g(t_0)|\,du=o(h^2)$. Thus $D_{f_k}(t_0,t_0+h)/D_w(t_0,t_0+h)\to f_k(t_0)/w(t_0)>c$. The proposed smaller constant fails for sufficiently small $h$.
\end{proof}

Sharpness is distribution-free. For a fixed dataset or restricted model family, predictions may never visit a region that attains the worst ratio. Consequently, the best distribution-free constant need not identify the best empirical predictor.

\section{The budgeted envelope loss}
Define $H(c)=\int_0^1(F(t)/c-m)_+\,dt$ for $c>0$, where $(u)_+=\max(u,0)$.

\begin{theorem}[Optimal allocation]\label{thm:design}
There is a unique $c_*>0$ satisfying $H(c_*)=B$. The weight
\begin{equation}\label{eq:optimal}
 \boxed{\quad w_*(t)=\max\{m,F(t)/c_*\}\quad}
\end{equation}
belongs to $\mathcal W_{m,B}$ and uniquely minimizes $C(w)$ up to almost-everywhere equality. Its optimal value is $C(w_*)=c_*$. Substitution of $w_*$ into \eqref{eq:loss} defines the envelope loss.
\end{theorem}
\begin{proof}
The envelope is bounded and integrable, with $1\leq\int F\leq K$. Dominated convergence on compact subintervals of $c>0$ makes $H$ continuous. It tends to infinity as $c\downarrow0$ and is zero when $c\geq\|F\|_\infty/m$. It is strictly decreasing wherever it is positive: for $c_1<c_2$ with $H(c_2)>0$, the positive-measure set $F>mc_2$ gives a strict decrease. Existence and uniqueness follow for $B>0$.

If $C(w)\leq c$, then $w\geq F/c$ and $w\geq m$ almost everywhere. Therefore $B=\int(w-m)\geq H(c)$, which forces $c\geq c_*$. The proposed $w_*$ attains the mass constraint. On the positive-measure active set $F>mc_*$ its ratio is $F/w_*=c_*$; elsewhere the ratio is at most $c_*$. It attains the lower bound. If another weight attained it, it would dominate $w_*$ pointwise almost everywhere and have the same integral. It must therefore equal $w_*$ almost everywhere.
\end{proof}

For $B=0$ the sole feasible weight is $m$ and the coefficient is $\|F\|_\infty/m$. For $B>0$, bisection or a bracketed scalar solver computes $c_*$. Feasibility also gives the useful lower bound
\begin{equation}
 c_*\geq\frac{\int_0^1F(t)\,dt}{m+B}.
\end{equation}
Equality holds if $F/c_*$ exceeds the floor almost everywhere. Regions below the floor threshold are left at $m$, while the other regions share a common density-to-weight ratio.

\paragraph{A closed-form example.}
Take $f_1(t)=2\1\{0<t<1/2\}$, $f_2(t)=4\1\{3/4<t<1\}$, $m=B=1$. The envelope is $2$ on the first interval, $4$ on the second, and zero elsewhere. The budget equation gives $c_*=8/7$. The optimal weights are $7/4$, $7/2$, and $1$ on these three regions, respectively. Uniform weight $2$ has coefficient $2$. The equally averaged weight $1+(f_1+f_2)/2$ has coefficient $4/3$. The optimized coefficient $8/7$ improves both, with identical total weight $2$.

\paragraph{What the optimum does not say.}
The training objective is ordinarily empirical $\ell_{w_*}$ loss. It is not $\max_k\mathcal R_{f_k}$ itself. Nor does minimizing $c_*$ minimize the product $c_*\mathcal R_{w_*}(q)$ over fitted models: changing $w$ changes approximation error, estimation error, and optimization. The theorem optimizes the multiplier for a fixed feasible loss scale. Log loss has weight $1/[t(1-t)]$, whose integral diverges, so it is outside \eqref{eq:feasible}; the theorem is not an optimality claim over log loss.

\section{Misspecification and limits}
\begin{proposition}[Errors in the threshold family]\label{prop:stability}
Suppose $\widehat f_k$ are bounded densities, $\widehat F=\max_k\widehat f_k$, and $\|F-\widehat F\|_\infty\leq\delta$. Let $\widehat w$ be the envelope design for $\widehat F$ under the same $m,B$, with coefficient $\widehat c$. Then
\begin{equation}
 C_F(\widehat w)\leq\widehat c+\delta/m,
 \qquad |c_*(F)-c_*(\widehat F)|\leq\delta/m.
\end{equation}
Here $C_F(w)=\esssup F/w$. Alternatively, if $\max_k\|f_k-\widehat f_k\|_1\leq\epsilon$, then for every predictor
\begin{equation}
 \max_k\mathcal R_{f_k}(q)\leq\widehat c\mathcal R_{\widehat w}(q)+\epsilon.
\end{equation}
\end{proposition}
\begin{proof}
For any feasible $w$, $|C_F(w)-C_{\widehat F}(w)|\leq\delta/m$. Apply this first to $\widehat w$, and then take infima in both directions to compare the optima. For the second result, the integrand $|t-\eta|\1\{t\text{ between }\eta\text{ and }q\}$ is bounded by one. The difference between the two conditional regrets is at most $\|f_k-\widehat f_k\|_1$. Average over $X$ and apply Theorem~\ref{thm:sharp} for the estimated family.
\end{proof}

These statements concern density approximation, not label noise, covariate shift, or a changing conditional event probability. Estimating the family from data requires a separate analysis of that estimation procedure. In the experiments below the laws are specified in advance, so there is no fitted-density uncertainty.

\begin{proposition}[An atomic threshold prevents linear domination]\label{prop:atom}
Let a threshold law place mass $a>0$ at $t_0\in(0,1)$. If $w$ is bounded by $M<\infty$ in a neighborhood of $t_0$, no finite constant $C$ bounds its decision regret by $CD_w(\eta,q)$ uniformly in $(\eta,q)$.
\end{proposition}
\begin{proof}
Choose $\eta=t_0-h$, $q=t_0+h$, and small $h>0$. The atom contributes $ah$ to decision regret. On the other hand, $D_w(\eta,q)\leq M\int_{\eta}^q(t-\eta)dt=2Mh^2$. Their ratio is at least $a/(2Mh)$ and diverges.
\end{proof}

The density assumption is therefore substantive. Smoothing a fixed threshold into a narrow density changes the target, even if it is operationally reasonable. The floor still gives the weaker fixed-threshold bound $\mathcal R_{t_0}(q)\leq\sqrt{2\mathcal R_w(q)/m}$: conditional threshold regret is at most $|q-\eta|$, followed by Cauchy--Schwarz and \eqref{eq:domination}.

\section{Implementation}
The loss is evaluated directly, not by multiplying a Brier loss by a prediction-dependent weight. In general $w(q)(y-q)^2$ has extra derivative terms and is not the proper loss in \eqref{eq:loss}. At differentiability points, a logit $z$ with $q=\sigma(z)$ gives
\begin{equation}\label{eq:gradient}
 \frac{\partial\ell_w(y,\sigma(z))}{\partial z}
       =(q-y)w(q)q(1-q).
\end{equation}
A bounded weight bounds this derivative by $\|w\|_\infty/4$, but also permits gradients to vanish near the endpoints. Strict properness does not imply convexity in the logit, so optimizer success alone is not a global-optimality certificate.

For histogram densities on $J$ bins with edges $0=e_0<\cdots<e_J=1$, form $F_j=\max_k f_{kj}$ and solve
\begin{equation}
 \sum_{j=1}^J(e_j-e_{j-1})(F_j/c-m)_+=B.
\end{equation}
Then $w_j=\max(m,F_j/c)$. Prefix integrals give exact partial losses. Define $A(q)=\int_0^qw(t)dt$ and $Q(q)=\int_0^qt w(t)dt$. If $q$ lies in bin $j$,
\begin{align}
 A(q)&=A(e_{j-1})+w_j(q-e_{j-1}),\\
 Q(q)&=Q(e_{j-1})+\tfrac12w_j(q^2-e_{j-1}^2).
\end{align}
Thus $\ell_w(0,q)=Q(q)$ and $\ell_w(1,q)=A(1)-Q(1)-A(q)+Q(q)$. Preprocessing costs $O(KJ)$ plus the scalar solve; each prediction needs a bin lookup and constant arithmetic. The integral is exact for the histogram laws, which are the actual experimental targets. At a knot the partial losses are continuous but may have unequal one-sided derivatives; the implementation uses the right bin.

\section{Numerical study}
The study has two purposes: verify the design calculation and examine whether coefficient improvements translate to learning performance. All results in this section were generated by the accompanying Python program. No proprietary or real-world data are used.

\subsection{Design verification}
\begin{figure}[htbp]
\centering\includegraphics[width=\linewidth]{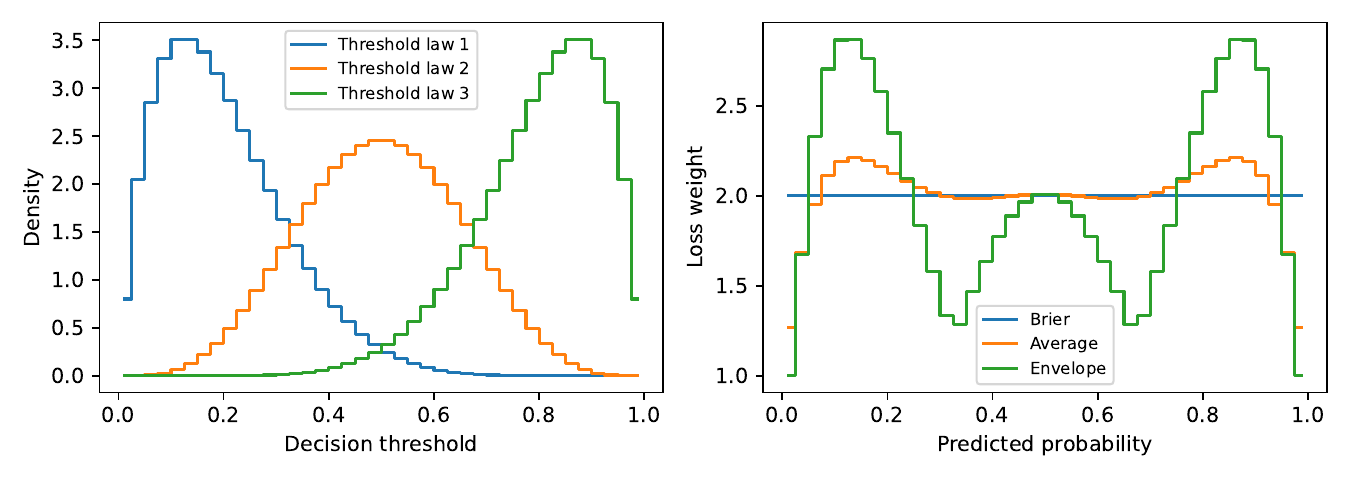}
\caption{Specified histogram threshold laws (left) and feasible loss weights (right). The envelope design allocates weight to the family maximum while respecting the floor and total mass.}\label{fig:weights}
\end{figure}

We use 40 equal-width bins and three densities obtained from the bin probabilities of $\mathrm{Beta}(2,8)$, $\mathrm{Beta}(5,5)$, and $\mathrm{Beta}(8,2)$. The bin probabilities are divided by bin width. We set $m=B=1$ and compare uniform $w=2$, average-density $w=1+\frac13\sum_k f_k$, and the envelope design (Figure~\ref{fig:weights}). Each weight integrates to two and satisfies the same floor.
\begin{table}[htbp]\centering
\caption{Sharp multipliers for the specified histogram family. Smaller is better at the fixed weight budget.}\label{tab:cert}
\begin{tabular}{lr}\toprule Weight & $C(w)$\\\midrule
Uniform (Brier) & 1.754774\\
Average density & 1.600765\\
Envelope & 1.223051\\\bottomrule\end{tabular}
\end{table}
The envelope coefficient is 30.30\% lower than the uniform coefficient and 23.60\% lower than the average-density coefficient. These are coefficient reductions, not reductions in prediction error. A separate linear program agrees with the scalar-root optimum within $9\times10^{-14}$. A local regret ratio in an active histogram bin agrees with it within $1.1\times10^{-11}$. A central finite-difference check of the probability derivative has maximum absolute error $1.5\times10^{-10}$ or less. Checks on 10,000 random probability pairs satisfy both the floor bound and each family domination bound within $10^{-12}$ numerical tolerance. Numerical checks supplement, rather than replace, the proofs.

\subsection{Learning protocol}
For each seed $s\in\{0,\ldots,19\}$, draw independent coordinates $X_1,X_2\sim\mathrm{Uniform}[-2,2]$. In the correctly specified setting,
\[
 \eta(X)=\sigma(0.3+1.1X_1-0.8X_2).
\]
In the misspecified setting, add $1.1\sin(2X_1)+0.7X_1X_2$ inside the sigmoid. Labels are independent Bernoulli draws with probability $\eta(X)$. Each setting uses 1,500 training observations and a separate 12,000-observation test sample per seed. Each method receives identical data within a setting and seed.

All methods fit $q_\theta(X)=\sigma(\theta_0+\theta_1X_1+\theta_2X_2)$. The four objectives are log loss, Brier loss, average-density proper loss, and envelope loss. All use the same $10^{-6}\|\theta_{1:2}\|_2^2/2$ penalty, with no intercept penalty. L-BFGS-B uses at most 1,000 iterations, function tolerance $10^{-12}$, and gradient tolerance $10^{-8}$. The starts are $(0,0,0)$, $(0.2,0.5,-0.5)$, and $(-0.2,-0.5,0.5)$; the lowest training objective selects the fit. All 160 selected fits report optimizer success. This does not certify global minimization for the three non-log objectives. No settings are selected on the test set.

The synthetic conditional probability is known. The primary metric is the maximum, over the three threshold laws, of the test-sample mean decision regrets in Equation~\eqref{eq:decision}. We also report $\E_X(q-\eta)^2$ and observed-label test log loss. These metrics use the same held-out samples; $\eta$ is never used for fitting. The reported errors are standard errors of the 20 seed-level estimates, not confidence intervals for real deployment performance.

\input{results/table.tex}

\subsection{Interpretation}
In the correctly specified setting, log loss has the lowest mean worst-family regret. In the misspecified setting, the average-density loss has a slightly smaller mean worst-family regret than the envelope loss; their seed-level uncertainties are larger than this small difference. Envelope loss improves on log loss for this decision metric under misspecification but has worse test log loss. These results do not establish a general advantage over Brier or average-density training. They illustrate the distinction predicted by the theory: optimizing a uniform regret multiplier is not equivalent to optimizing the error of a fitted, restricted model.

Only two synthetic data-generating processes and one linear model family are tested. There is no evidence here about neural networks, large-scale auctions, medical decisions, label imbalance, or real-world deployment. The settings were chosen to contrast correct specification with a fixed nonlinear misspecification, not to approximate any particular production system.

\section{Discussion}
The envelope construction gives a reproducible way to select one proper loss for a finite family of threshold preferences under an explicit scale convention. Its precise benefit is a minimax, distribution-free regret multiplier. The optimization is elementary once the proper-score representation is available; the technical content lies in stating the feasible class, proving the coefficient is sharp, and identifying its unique optimum.

Several issues remain outside this result. A practitioner must justify the threshold family and floor-to-budget ratio. A larger budget changes the loss scale and its allocation; coefficients at different budgets should not be read as an empirical comparison. A feature-dependent threshold law would require a conditional formulation. Atomic thresholds need a different guarantee or a measure-valued loss. Finite-sample excess-risk bounds, optimal regularization, and tests on real data remain open extensions of this study. Most immediately, the empirical results caution against selecting a training loss solely from its worst-case multiplier.

\paragraph{Reproducibility and assistance.}
The accompanying source package contains the experiment program, per-seed results, figure, and verification code. Computation used Python, NumPy, SciPy, and Matplotlib on CPU. OpenAI ChatGPT/Codex assisted with the mathematical development, literature search, manuscript drafting, and experiment implementation and execution. This statement describes the assistance used; it does not imply independent human verification or peer review.

\end{document}

%% file: results/table.tex
\begin{table}[htbp]\centering
\caption{Test performance, mean $\pm$ standard error over 20 seeds. Worst-family regret and probability MSE are multiplied by $10^3$. Lower is better.}
\label{tab:learning}
\small
\begin{tabular}{llrrr}\toprule
Setting & Loss & Worst regret $\times10^3$ & MSE $\times10^3$ & Log loss\\\midrule
Correct & Log & $0.268 \pm 0.059$ & $0.351 \pm 0.080$ & $0.505 \pm 0.001$\\
Correct & Brier & $0.324 \pm 0.066$ & $0.407 \pm 0.086$ & $0.505 \pm 0.001$\\
Correct & Average & $0.321 \pm 0.066$ & $0.404 \pm 0.086$ & $0.505 \pm 0.001$\\
Correct & Envelope & $0.302 \pm 0.069$ & $0.382 \pm 0.088$ & $0.505 \pm 0.001$\\
Misspecified & Log & $17.695 \pm 0.108$ & $26.352 \pm 0.151$ & $0.493 \pm 0.001$\\
Misspecified & Brier & $16.965 \pm 0.100$ & $25.316 \pm 0.121$ & $0.500 \pm 0.002$\\
Misspecified & Average & $16.955 \pm 0.100$ & $25.314 \pm 0.118$ & $0.501 \pm 0.002$\\
Misspecified & Envelope & $16.978 \pm 0.098$ & $25.305 \pm 0.109$ & $0.501 \pm 0.002$\\
\bottomrule\end{tabular}\end{table}